%% file: main.tex
\documentclass[sigconf]{acmart}
\usepackage{hyperref}
\usepackage{tcolorbox}
\usepackage{multirow}
\usepackage{algorithm}
\usepackage{algorithmic}
\usepackage{subcaption} 

\AtBeginDocument{%
  }

\setcopyright{acmlicensed}
\copyrightyear{2025}
\acmYear{2025}
\acmDOI{XXXXXXX.XXXXXXX}

\acmConference[WWW '25]{the ACM Web Conference 2025}{April 28--May 05, 2025}{Sydney, Australia}

\acmISBN{978-1-4503-XXXX-X/18/06}

\begin{document}
\settopmatter{authorsperrow=4}
\title{LLM-powered Multi-agent Framework for Goal-oriented Learning in Intelligent Tutoring System}



\author{Tianfu Wang}
\affiliation{
  \institution{
  HKUST (GZ)
  }
  \city{Guangzhou}
  \country{China}
}
\email{tianfuwang.cs@outlook.com}

\author{Yi Zhan}
\affiliation{%
  \institution{Microsoft Inc.}
  \country{China}
}
\email{yzhan0119@outlook.com}

\author{Jianxun Lian}
\affiliation{%
  \institution{Microsoft Research Asia}
  \city{Beijing}
  \country{China}
}
\email{jianxun.lian@outlook.com}

\author{Zhengyu Hu}
\affiliation{%
  \institution{HKUST (GZ)}
  \city{Guangzhou}
  \country{China}
}
\email{zhu021@connect.hkust-gz.edu.cn}

\author{Nicholas Jing Yuan}
\authornote{Corresponding authors.}
\affiliation{%
  \institution{Microsoft Inc.}
  \country{China}
}
\email{nicholas.yuan@microsoft.com}

\author{Qi Zhang}
\affiliation{%
  \institution{Microsoft Inc.}
  \country{China}
}
\email{zhang.qi@microsoft.com}

\author{Xing Xie}
\affiliation{%
  \institution{Microsoft Research Asia}
  \city{Beijing}
  \country{China}
}
\email{xingx@microsoft.com}

\author{Hui Xiong}
\authornotemark[1]
\affiliation{%
  \institution{HKUST (GZ)}
  \city{Guangzhou}
  \country{China}
}
\email{xionghui@ust.hk}

\renewcommand{\shortauthors}{Wang et al.}

\input{sections/abstract}

\ccsdesc[500]{Information systems~Web mining}

\renewcommand\footnotetextcopyrightpermission[1]{}
\settopmatter{printacmref=false}
\keywords{Intelligent Tutoring System, Large Language Model, Multi-agent}


\input{sections/abstract}
\maketitle

\input{sections/introduction}

\input{sections/related-work}

\input{sections/methodology}

\input{sections/evaluation}

\input{sections/conclusion}

\clearpage
\bibliography{ref}
\bibliographystyle{plain}

\input{sections/appendix}

\end{document}

%% file: sections/abstract.tex
\begin{abstract}
Intelligent Tutoring Systems (ITSs) have revolutionized education by offering personalized learning experiences. However, as goal-oriented learning, which emphasizes efficiently achieving specific objectives, becomes increasingly important in professional contexts, existing ITSs often struggle to deliver this type of targeted learning experience. In this paper, we propose \textbf{GenMentor}, an LLM-powered multi-agent framework designed to deliver goal-oriented, personalized learning within ITS. GenMentor begins by accurately mapping learners' goals to required skills using a fine-tuned LLM trained on a custom goal-to-skill dataset. After identifying the skill gap, it schedules an efficient learning path using an evolving optimization approach, driven by a comprehensive and dynamic profile of learners' multifaceted status. Additionally, GenMentor tailors learning content with an exploration-drafting-integration mechanism to align with individual learner needs. Extensive automated and human evaluations demonstrate GenMentor's effectiveness in learning guidance and content quality. Furthermore, we have deployed it in practice and also implemented it as an application. Practical human study with professional learners further highlights its effectiveness in goal alignment and resource targeting, leading to enhanced personalization. Supplementary resources are available at \href{https://github.com/GeminiLight/gen-mentor}{https://github.com/GeminiLight/gen-mentor}.
\end{abstract}

%% file: sections/introduction.tex
\section{Introduction}
Intelligent Tutoring Systems (ITSs) have made significant strides in supporting personalized learning experiences by leveraging machine learning (ML) technologies~\cite{its-arr-1990-overview}. These systems increasingly utilize data-driven insights across key aspects, such as managing learning materials~\cite{kdd-2019-learning-path,kbs-2023-learning-path-rl}, profiling learner status~\cite{kdd-2022-cognitive-diagnosis,kt-www-2024-interpretable}, and delivering personalized feedback~\cite{edu-www-2019-learnerexp,sigir-2022-education-question-retrieval}. However, many ML-based ITSs suffer from fragmentation due to technical inconsistencies across modules. They also exhibit limited adaptability to emerging topics, often requiring extensive retraining to accommodate new educational topics. Recently, large language models (LLMs) have presented new opportunities to enhance ITS by offering interactive feedback through conversational interfaces~\cite{its-aied-2023-dialog-based-tutoring}. Existing studies on LLM-based ITS~\cite{its-chi-2024-conversation,its-cikm-2024-private-tutoring,its-arxiv-2024-personality} demonstrate their facilitation in content generation, query response, and problem-solving. However, while LLMs excel at dialogue-based engagement, they often remain reactive, merely responding to learner queries without proactively guiding learners toward their objectives~\cite{its-eacl-2023-opportunities-neural-dialog-tutoring}. This reactivity also limits their capacity to gain a comprehensive understanding of learners, thus weakening the personalization~\cite{its-chi-2024-conversation}.

As educational demands continue to diversify, especially within professional and lifelong learning contexts, learners increasingly seek systems that support personal or career-specific goals~\cite{its-book-2020-lifelong-learning}. For example, an employee assigned a task or seeking new jobs that involve unfamiliar technology may feel uncertain about where to start or what specific skills to develop. 
Without clear guidance, they risk becoming overwhelmed or wasting time on less relevant content~\cite{its-val-2022-goal-orientation}.
Such learners would benefit from a system that bridges this gap by quickly identifying their knowledge deficiencies and providing targeted content to help them acquire the precise skills needed to complete the objective efficiently.
This personalized approach, known as goal-oriented learning, goes beyond merely delivering information, focusing on guiding goal achievement~\cite{its-ce-2021-goal-oriented-active-learning}.
However, traditional ITSs often rely on static curricula, offering broad content that may not adequately address the unique needs of individual learners.
Additionally, LLM-based dialogue ITSs tend to be reactive rather than actively guide learners to achieve goals.
As a result, these systems struggle to provide goal-oriented, personalized guidance, especially in fast-paced, professional contexts.

\begin{figure}
    \centering
    \includegraphics[width=0.478\textwidth]{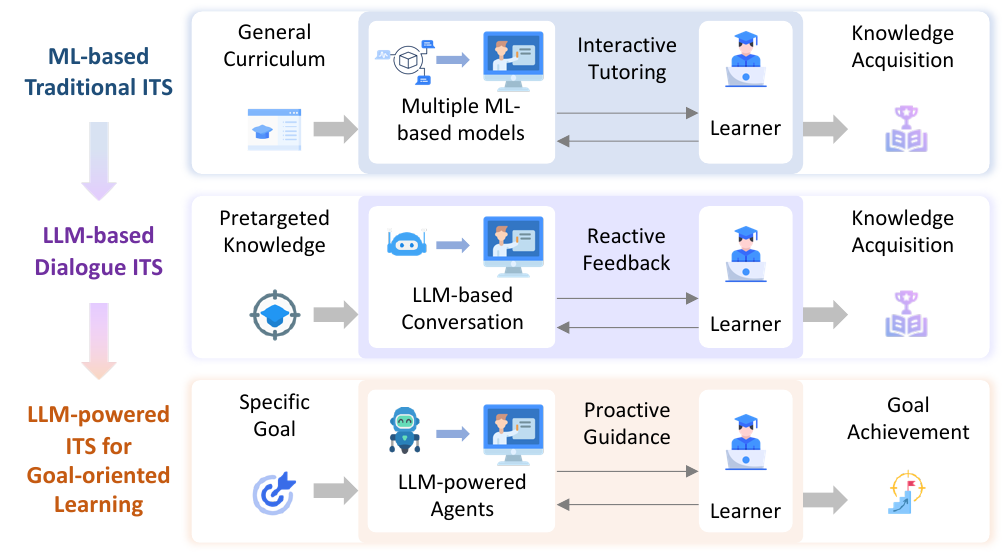}
    \vspace{-10pt}
    \caption{Comparison of three types of ITS Paradigms.}
    \label{fig:paradigm-comparison}
    \vspace{-16pt}
\end{figure}

To address these gaps, we propose a perspective shift toward goal-oriented learning for transforming ITSs. 
As illustrated in Figure~\ref{fig:paradigm-comparison}, unlike traditional ITSs that rely primarily on static curricula or reactive engagement, our approach emphasizes proactively guiding learners to achieve their specific goals. 
In goal-oriented ITSs, the learning process is driven by the learner's objectives rather than a predefined curriculum. 
By identifying skill gaps and profiling learner status, these systems can deliver more personalized pathways and tailored content, enabling learners to rapidly acquire the necessary skills to meet their specific goals. 
Notably, LLMs hold promising potential due to their remarkable ability to understand complex intentions~\cite{llm-emnlp-2024-intent} and generate versatile content, making them well-suited for goal-oriented learning experiences~\cite{llm4edu-arxiv-2024-survey-outlook}.

Despite these benefits, several challenges should be addressed to ensure the effective functioning of LLM-powered goal-oriented ITSs.  
First, accurately identifying the learner's skill gaps in relation to their specific goals is critical, i.e., mapping the learner's objectives to the newly required skills for progress.
This necessitates that LLMs handle a nuanced understanding of both the learner's profile and the target goal.
Second, robust learner modeling is crucial for capturing individual preferences, progress, and cognitive abilities. The system must continuously adapt to real-time feedback, ensuring the learning experience evolves with the learner's needs and abilities, while dynamically adjusting strategies based on the learner's state.
Third, providing personalized resources, e.g., both learning paths and delivered content, facilitates efficient knowledge acquisition. The system must ensure that resources are relevant, goal-aligned, and adaptable to learner progress, presenting challenges in content curation and learning path optimization.

In this work, we introduce \textbf{GenMentor}, an LLM-powered multi-agent ITS designed to deliver a personalized, goal-oriented learning experience.
To address the diverse functions required in ITSs, we distribute responsibilities across multiple LLM agents, allowing them to collaboratively manage distinct tasks~\cite{llm-ijcai-2024-multiagents}. Specifically, to improve the relevance and goal alignment of identified skills, we customize a goal-to-skill dataset to fine-tune the LLM as \textit{skill identifier}, which contributes to accurately mapping learning goals to the necessary skills. Then, this agent identifies the skill gap by additionally considering learner's initial information.
To enable learner-specific customization, we employ an adaptive \textit{learner profiler} to dynamically track the learner's cognitive status, preferences, and behavioral patterns. By integrating real-time interactions, it also continuously updates the learner's profile for a deeper and more accurate understanding.
Furthermore, to close the identified skill gap, the \textit{path scheduler} continuously refines the learning path for effective goal achievement, where a \textit{learner simulator} with generated profile is used to mimic learner feedback on provided resources.
We also implement an \textit{content creator} with an exploration-generation-integration mechanism, i.e., first exploring goal-oriented knowledge and preparing an outline, then drafting each section, and finally integrating and refining them to create learning materials. 
This process ensures that delivered content remains aligned with the learner's goals while enhancing personalization and effectiveness.

To evaluate the effectiveness of GenMentor, we conduct both automated and human evaluations, which show its superiority in identified skill requirements, scheduled learning paths, and generated content. 
Additionally, we have deployed it in the practical product and implemented an independent application designed for professional employee learning and goal achievement. A human study further highlights its effectiveness in providing learning guidance and targeting resources, and facilitating goal achievement.

We summarize the main contributions of our work as follows:
\begin{itemize}
    \item We introduce a novel perspective on ITS, goal-oriented learning, to meet the practical educational needs of learners. Through the integration of LLMs, we propose GenMentor, an LLM multi-agent system to implement this philosophy.
    
    \item To improve goal alignment of identified skills, we build a goal-to-skill mapping dataset for LLM fine-tuning. Grounded in educational theories, we propose a adaptive learner modeling method to capture evolving needs. We design an evolvable learning path scheduling and an exploration-draft-integration method to enhance resource personalization.

    \item We conduct extensive experiments through both automated and human evaluation. The results demonstrates that GenMentor offers superior learning guidance and resources.

    \item We deploy GenMentor in a practical product and implement it as an application. A study with professional learners also shows its effectiveness in achieving goal-oriented learning.
\end{itemize}

%% file: sections/related-work.tex
\section{Related Works}

\subsection{Intelligent Tutoring System}
Intelligent Tutoring Systems (ITSs) are designed to provide personalized experiences, mimicking one-on-one human tutoring~\cite{its-arxiv-2018-survey}.
With advances in machine learning, modern ITSs have shifted toward leveraging data-driven approaches for material management~\cite{kdd-2019-learning-path,kbs-2023-learning-path-rl}, learner modeling~\cite{kdd-2022-cognitive-diagnosis,kt-www-2024-interpretable}, and feedback provision~\cite{edu-www-2019-learnerexp,sigir-2022-education-question-retrieval}.
Despite these advancements, existing ML-based ITSs often combine several ML-based models for distinct tasks, resulting in technical inconsistencies and data fragmentation, and lacking generalizability for emerging topics. Additionally, they rely on static content and predefined curricula, focusing on general knowledge acquisition rather than dynamically adjusting to individual learner goals.
Recently, several works~\cite{its-chi-2024-conversation,its-cikm-2024-private-tutoring,its-arxiv-2024-personality,edu-arxiv-2024-tutoring-dataset} have integrated LLMs into ITS as a conversational tutor due to their excellent abilities in understanding and generation of LLMs. Otherwise, these systems can not proactively guide learners toward their long-term objectives due to their reactive nature and dialogue-based engagement. 
In our work, we explore a unified LLM-based multi-agent framework for ITS, which addresses the fragmentation that often arises from separate traditional ML models for ITS and enhances consistency across tutoring phases. Furthermore, we focus on providing proactive learning guidance for goal achievement, which differs from traditional ITS for broad knowledge acquisition.

\subsection{LLM-powered Multi-agent Systems}
Unlike individual LLM agents that operate independently, LLM-based multi-agent systems employ multiple LLM agents with diverse profiles to collaboratively tackle tasks~\cite{llm-ijcai-2024-multiagents}. By distributing responsibilities, these systems can address more dynamic and complex challenges. This approach has shown promising results across various domains, such as software development~\cite{llm-acl-2024-chatdev}, 
embodied agent~\cite{llm-icra-2024-roco} and game playing~\cite{llm-arxiv-2024-werewolf}. For instance, in software development, a chat-powered framework was proposed where LLM agents collaborate across different phases, (i.e., design, coding, and testing), through unified language-based communication. Additionally, LLM multi-agents have been explored in society simulation~\cite{llm-uist-2023-simulacra,llm-arxiv-2023-metaagents}, where they simulate human behaviors and provide feedback, mimicking real-world interactions. These studies highlight the human-like capabilities of LLM agents and offer valuable insights into human behavior and decision-making.
In the context of ITSs, they involve complex and evolving tasks such as personalized content generation, adaptive learning path design, and real-time student feedback processing.
To handle such intricacies and enhance overall consistency, we leverage the collaborative power of LLM agents in ITS stems from their ability to handle multiple specialized roles, enabling more personalized and adaptive learning experiences.

%% file: sections/methodology.tex
\begin{figure*}[htp]
    \centering
    \includegraphics[width=0.942\textwidth]{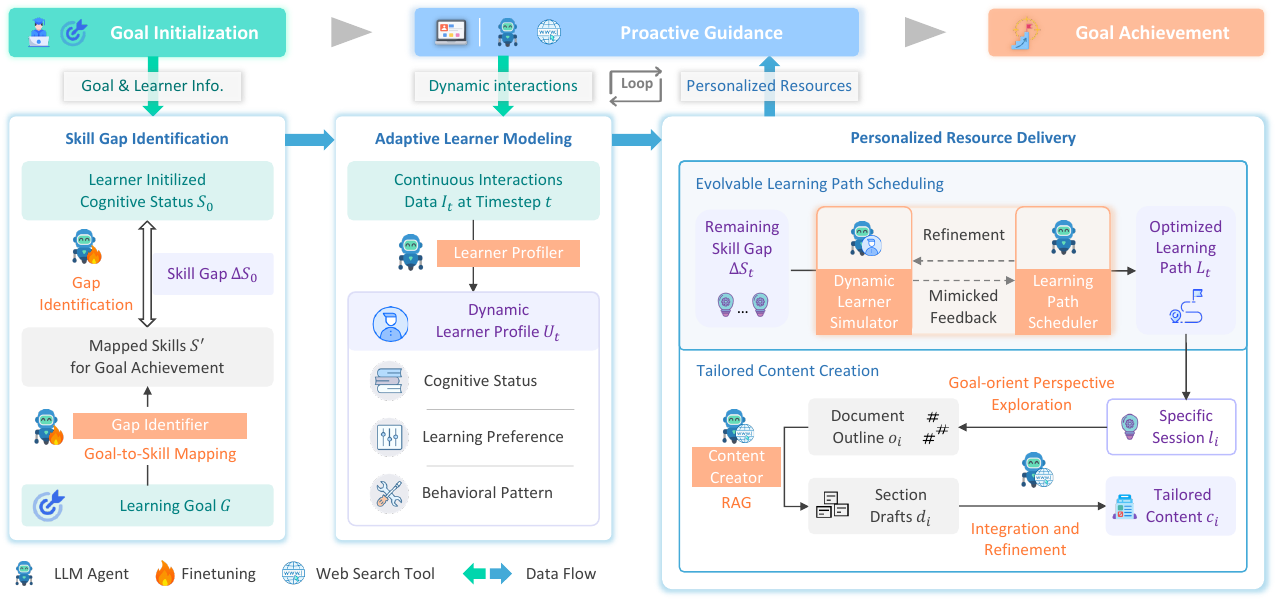}
    \vspace{-10pt}
    \caption{Overview of the GenMentor: An LLM-powered multi-agent framework for goal-oriented learning in ITS.}
    \label{fig:llm4edu-framework}
    \vspace{-10pt}
\end{figure*}

\section{Problem Statement}

In goal-oriented learning, the focus is on achieving specific objectives efficiently, such as completing a project or mastering job-related skills. This requires proactive guidance to avoid unfocused learning and to improve goal completion efficiency. A goal-oriented ITS aims to customize learning pathways and content, enabling learners to quickly acquire the knowledge needed to meet their specific goals.
Formally, let $U_0 = (S_0, P, B)$ represent the learner's profile, where $S_0, P, B$ denote the learner's initial knowledge status, preferences, and behavior patterns, respectively. 
Given a specific goal $G$, the learner must master a set of skills $S^\prime$ necessary to achieve this objective.
Our aim is to efficiently minimize the skill gap $\Delta S_0 = S^\prime - S_0$ with a personalized and adaptive learning experience by scheduling a personalized learning path $L$ and tailored contents $C$ based on $P$, $B$, and real-time learner interactions $I$. To accomplish this, the system should address three key sub-tasks:

\begin{itemize}
    \item \textit{Skill Gap Identification}. This step identifies the skills gap between the learners' current knowledge and the skills required to achieve their goals. First, goal $G$ is mapped to the necessary skills, $f: G \rightarrow S^\prime$, and then the skill gap $\Delta S_0$ is identified by $f: (S_0, S^\prime) \rightarrow \Delta S_0$, where the $S_0$ is derived from learner's provided individual information $I_0$.
    \item \textit{Adaptive Learner Modeling}. This module continuously update learner profiles by incorporating interaction data $I_t$ at timestep $t$, yielding $f: (U_{t-1}, \Delta S_0, I_t) \rightarrow (U_t, \Delta S_t)$. This enables the system to track cognitive progress, recognize learning preferences, identify and behavior patterns.
    \item \textit{Personalized Resource Delivery}. To efficiently close the skill gap $\Delta S_t$, this module dynamically schedules an engaging learning path $L_t$ and delivers tailored learning content $C_t$, i.e., $f: (U_t, \Delta S_t) \rightarrow (L_t, C_t)$.
    The path should adapt to the learner's evolvable progress and preferences, while the content should be high-quality, goal-relevant and personalized.
\end{itemize}

\section{The GenMentor Framework}
In this work, we propose an LLM-powered multi-agent framework for goal-oriented learning in ITS, named GenMentor. To address the above intricate sub-tasks, we distribute responsibilities across multiple LLM agents, allowing them to collaboratively manage different tasks.
As illustrated in Figure~\ref{fig:llm4edu-framework}, this system begins with accurately assessing the skill gap between the learner's current skills and the target objective. Once this gap is identified, the ITS generates a personalized learning path with an evolvable optimization method for both learning efficiency and engagement.
Furthermore, the system curates and generates content that is goal-oriented, up-to-date, and tailored to the learner's specific needs, ensuring the learner focuses on the most relevant and targeted content. 
During the learning process, the generated learner profile are continuously adjusted with the newly learner's interactions, enabling dynamical adaptation to the learner's evolving progress, preferences and needs. 

\subsection{Skill Gap Identification}
To personalize the learning experience toward achieving specific goals, we identify the skill gap, i.e., the necessary skills bridging the learner's current cognitive status and target objectives.
Mapping these goals accurately requires the LLM to grasp the goal's nuances, given their often abstract and high-level nature.
However, direct prompting LLMs may produce irrelevant, unnecessary or incomplete skills, impeding effective goal achievement.
Thus, we build a customized goal-to-skills dataset with Chain of Thought (CoT)~\cite{llm-neurips-2022-cot} to fine-tune LLMs, improving the goal alignment of identified gap.

\subsubsection{CoT-enabled Dataset Construction}
Given the absence of a directly relevant dataset for our task, we turn to job posting datasets, which contain detailed information on job roles, descriptions, and required skills. These datasets provide foundational insights into the expectations and requirements for various roles. 
By extracting pairs of job summaries (comprising roles and brief job descriptions) and their corresponding core skills, we construct a goal-to-skill dataset, treating these job summaries as goals.
However, direct fine-tuning on these datasets may fall short in accurately mapping goals to skills, largely due to the abstract and high-level language common in job descriptions. To address this limitation, we employ CoT reasoning~\cite{llm-neurips-2022-cot}, introducing intermediate steps that clarify the logical connections between job responsibilities and the necessary skills. The CoT process involves breaking down the goals into key tasks, identifying the required skills for each task, and determining the proficiency levels needed, producing samples of <job summary, reasoning tracks, required skills>. See Appendix~\ref{dataset-details} for more details.
This approach facilitates capturing the nuanced relationships between goals and skills, thereby improving fine-tuning accuracy.

\subsubsection{Fine-tuning LLM for Goal Alignment}
Using the constructed goal-to-skills dataset, we fine-tune the LLM as \textit{skill identifier} to accurately map goals to specific skills. This step ensures that this agent identifies relevant and complete skills while filtering out unnecessary skills, focusing on efficient goal achievement.

\subsubsection{Gap Identification Process}
This process begins with goal-to-skills mapping, $S^\prime = LLM_{\textit{skill-identifier}}(G)$, where the learning goal is mapped to a set of skills that are the core competencies needed for the goal. Next, identified skills are compared with learner's initial cognitive status based on provided information to establish the skill gap, $\Delta S = LLM_{\textit{skill-identifier}}(S^\prime, S_0)$. This step filters out already-mastered skills and highlights areas needing improvement.

\subsection{Adaptive Learner Modeling}
Understanding the learner's status is essential for ITS as it personalizes learning by adapting content to their needs, and providing targeted feedback~\cite{leaernermodelling-etrd-2019-systematic-review}. Instead of traditional ML-based methods lacking generalization and integration while LLM-based dialogue remain reactive, here, we explore how to explicitly leverage LLM to achieve the comprehensive and dynamic learner profile $U$.

\subsubsection{Comprehensive Learner Profile}

To capture learner's knowledge status and provide customized learning resources that align with goals and preferences, we consider three fundamental aspects to create a comprehensive learning profile $U$, informed by educational theories~\cite{leaernermodelling-book-2013-design-recommendations,leaernermodelling-etrd-2019-systematic-review,learnermodeling-ijmecs-2016-learner-modeling}. These aspects are as follows:

\begin{itemize}
    \item \textit{Cognitive Status} $S$. To monitor the learner's knowledge acquisition, we track learning progress and assess mastery of required skills. We represent these skills as a set of competencies the learner has mastered and those they still need to acquire, along with corresponding metrics of progress and mastery. This approach highlights remaining skill gaps and enables the system to provide targeted content to help the learner achieve their goals.

    \item \textit{Learning Preferences} $P$. Recognizing the diverse ways learners absorb information, this aspect captures individual preferences such as preferred content styles (e.g., concise summaries, detailed explanations) and preferred activity types (e.g., reading, active querying, interactive exercises).
    These insights enable the system to adapt its instructional methods and dynamically adjust the content delivery to enhance learner engagement and knowledge comprehension.

    \item \textit{Behavioral Patterns} $B$. By analyzing interaction data, we aim to identify behavioral trends that affect learning engagement, such the system usage frequency and the time consumption variability of learning sessions. Infrequent use and irregular time consumption (e.g., long time spent in one session) may signal disengagement or difficulty.
    These insights allow for proactive interventions, such as motivational messages or adjusted content difficulty, to maintain learner momentum toward their goals.
    
\end{itemize}

\subsubsection{Dynamic Learner Modeling}
Initially, based on provided information of learner $I_0$ (e.g., resume) and identified skill gaps $\Delta$, we use a \textit{learner profiler} to create a preliminary profile, i.e., $U_0 = LLM_{\textit{learner-profiler}}(I_0, \Delta S_0)$, capturing aforementioned three aspects.
This initial profile serves as the starting point, may having limitations or inaccuracies.
As the learner interacts with the system, their profile is continuously updated and refined based on collected dynamic interactions $I_t$ (e.g., real-time performance data and proactive feedback) at each timestep $t$, i.e., $U_t = LLM_{\textit{learner-profiler}}(U_{t-1}, I_t)$. 

\begin{figure}
    \centering
    \includegraphics[width=0.475\textwidth]{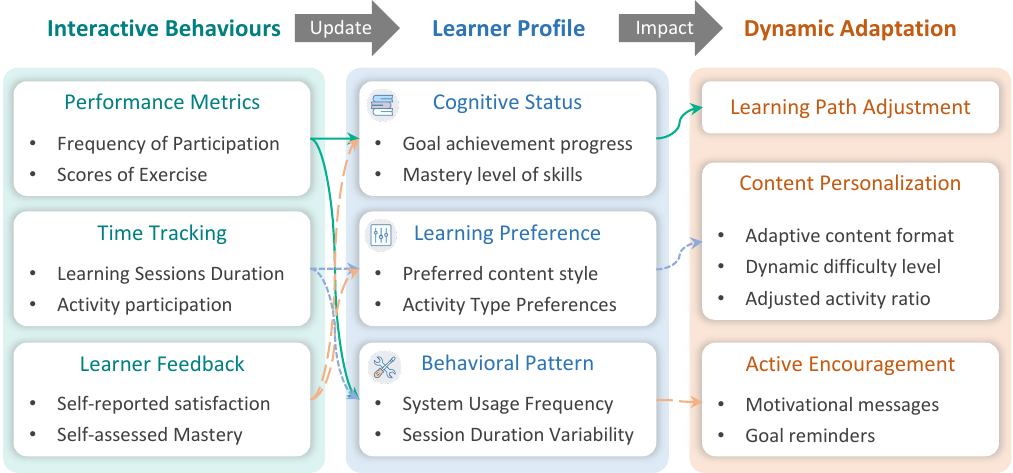}
    \vspace{-19pt}
    \caption{An illustration of dynamic learner modeling.}
    \vspace{-16pt}
    \label{fig:learner-modeling}
\end{figure}

During each session learning at timestep $t$, the system collects learner feedback and tracks metrics like performance and time use as interaction data $I_t$. For \textit{learning preferences} $P$, data on time spent on different activities and feedback identify favored content, enabling dynamic adjustment of instructional content. For example, if a learner favors interactive exercises, the system prioritizes these activities by offering more exercises. Regarding \textit{behavioral patterns} $B$, platform usage frequency and engagement consistency help gauge motivation. For example, irregular patterns (e.g., prolonged durations on single sessions or infrequent login) prompt interventions like trigger motivational prompts or adjusted content difficulty to maintain engagement and prevent frustration.
After each session, the system assesses the progress of \textit{cognitive status} by quiz scores and learner-reported feedback. It updates the learning process and skill mastery and identifies the remaining skill gap for future learning.
An illustrative example is shown in Figure~\ref{fig:learner-modeling}.

\subsection{Personalized Resource Delivery}
To effectively close identified skill gaps, GenMentor employs a personalized, adaptive content delivery method that dynamically aligns resources with each learner's unique profile and progress. Concretely, given the skill gap $\Delta S_0$ and learner profile $U_0$ obtained in the previous stage, the \textit{path schedule} creates a learning path $L = \{l_1, l_2, \cdots, l_n\}$, consisting of multiple learning sessions $l_i$.
For each learning session $l_i$, \textit{content creator} tailor content $C_i$ with an exploration-drafting-integration mechanism to improve the resource personalization and targeting. During the learning process, a \textit{ learner simulator} using the dynamic learner profile $U_t$ is employed to mimic the learner feedback on learning resources for refinement, enabling the adaptation of the learning path and content.

\subsubsection{Learner Simulator via Adaptive Profiling}
To achieve adaptations of learning resources while minimizing reliance on direct user feedback, we design an LLM agent-based feedback mechanism that simulates the learner responses. Specifically, leveraging real-time learner profiles $U_t$, a \textit{learner simulator} employs the role-playing method to anticipate learner feedback for delivered resources~\cite{llm-nature-2023-roleplay}. This simulation serves as a proxy to optimize delivered resources without requiring direct learner feedback.
For learning paths, the \textit{learner simulator} evaluates factors such as efficiency, engagement, and task difficulty, while anticipating learner reactions to tailored content (e.g., asking for more exercise, or feedbacking too high difficulties).  
This method allows the system to proactively adjust the learning resource to better match learner intentions and preferences, maximizing comprehension and sustaining motivation.

\subsubsection{Evolvable Learning Path Scheduling} We use a \textit{path scheduler} equipped with CoT reasoning ability to plan effective and engaged learning pathways that effectively support skill acquisition.
Initially, it constructs an initial learning path $L_0 = \text{LLM}_{\textit{path-scheduler}}(U_0, \Delta S_0)$ based on the identified skill gap $\Delta S_0$ and the initialized learner profile $U_0$. This path is refined iteratively with feedback from the learner simulator, ensuring the learning path becomes progressively challenging and motivational for the learner.
As the learning process progresses, the learner profile $U_t$ is continuously updated to incorporate insights into the learner's evolving preferences, abilities, and progress.
After completing a session at timestep $t$, the path scheduler dynamically re-evaluates and adjusts the learning path $L_{t-1}$ with the updated profile, while generating $L_t = \text{LLM}{\textit{path-scheduler}}(U_t, \Delta S_t, L_{t-1})$. If the learner approves, the updated learning path $L_t$ replaces the previous one.
By iteratively evaluating and adjusting the learning path, this adaptive scheduling mechanism ensures alignment with learner's evolving needs.

\subsubsection{Tailored Content Curation}
To offer learners relevant, personalized learning content, GenMentor employs a \textit{content creator} with an exploration-drafting-integration mechanism to curate materials, including documents and quizzes. To ensure accuracy and up-to-date information, the content creator integrates web search tools for retrieval-augmented generation (RAG). 
Specifically, 
for a given learning session $l_i$, the \textit{content creator} begins with the exploration to identify goal-related knowledge points across diverse perspectives. It then drafts content based on a systematic document outline considering the learner's preferences, and finally integrates feedback from the learner simulator to refine and finalize the learning material. This approach ensures that the content remains accurate and learner-focused, facilitating goal achievement.

\textit{Goal-oriented Knowledge Exploration}.
Key knowledge points are identified to ensure the learning content is both comprehensive and aligned with the learner's goals. Guided by practical learning~\cite{edu-book-2017-complex, edu-clt-2011-cognitive}, the content covers foundational concepts that provide the necessary background knowledge, practical insights that help bridge the gap between theoretical knowledge and real-world application, and problem-solving strategies that equip learners with techniques to address challenges, fostering critical thinking and adaptability. The \textit{content creator} uses the session title as a query, to retrieve the latest information. Combining this up-to-date information with the inherent knowledge of the LLM, it identifies the relevant knowledge points and organizes them into a structured document outline. This outline, derived by $o_i = \text{LLM}_{\textit{content-creator}}(U_t, \Delta S_t, L_t, l_i)$, serves as the foundation for subsequent drafting, ensuring that the content remains targeted, current, and aligned with the learner's objectives.

\textit{RAG-based Section Drafting}.
When drafting content of each section, to mitigate common issues in direct LLM generation such as hallucinations and long-tail inaccuracies~\cite{llm-acl-2023-hallucinations, llm-icml-2023-long-tail}, the content creator integrates high-quality retrieved information.
It formulates queries by combining the session title and section titles to retrieve relevant and reliable data. 
This retrieved information is then re-customized to align with the learner's profile, ensuring the content is both informative and engaging. These section drafts, $d_i = LLM_{\textit{content-creator}}(U_t, \Delta S_t, L_t, l_i, o_i)$, are tailored to match the learner's preference and progress, improving the personalization.

\textit{Integration and Refinement}.
The section drafts are synthesized into a cohesive document and then the learner simulator provides the mimicked feedback of learner and assesses logical structure and coherence. 
The content creator refines sections requiring improvement, ensuring each part aligns with the learner's needs. Once refined, these sections are seamlessly integrated into the final learning document.
Quizzes are generated alongside the document to enhance learner engagement and test knowledge mastery. This process is denoted as $c_i = LLM_{\textit{content-creator}}(U_t, \Delta S_t, L_t, l_i, o_i, d_i)$.
This method further ensures the final content is logically organized and learner-tailored, promoting both comprehension and motivation.

%% file: sections/evaluation.tex
\section{Experiments}
To evaluate the effectiveness of GenMentor's output items, we conduct both LLM-based automated and human evaluations.

\subsection{Implementations and Settings}
We implement GenMentor with two popular LLMs, GPT-4o (2024-08-06)~\cite{llm-arxiv-2023-gpt} and Llama 3.2~\cite{llm-arxiv-2023-llama} (3B). For the \textit{skill identifier}, both two LLMs are fine-tuned in Azure AI Studio on our custom goal-to-skill dataset, using a batch size of 3, and a maximum of 10 epochs. \textit{content creator} uses the Bing search tool to access the internet, retrieving up to 5 results per query to ensure concise and relevant information. In the RAG module, the text embeddings are generated by the text-embedding-3-small model. We set the LLM temperature of 0.7 across all experiments to balance consistency and creativity. 

\subsection{LLM-based Automated Evaluation}

Following prior works~\cite{llm-emnlp-2024-prometheus2,lm-naacl-2024-wikipedia,llm-neurips-2023-judging}, we use GPT4o as an automated evaluator due to its strong alignment with human judgments. We adopt the 5-point Likert scale assessment to evaluate outputs.

\subsubsection{Overall Experiment Setup}
We adopt a resume dataset~\footnote{https://www.kaggle.com/datasets/gauravduttakiit/resume-dataset} to represent diverse learner information and a subset of a job posting dataset to define target learning goals (excluded from the skill identifier model's training data). To construct the testing dataset, we randomly matched resumes with job postings to simulate the input information provided by learners and predefined goals. The testing dataset included five types of occupations, with a total of 200 samples. For each sample, we simulate a learner with a specific resume as learner information $S_0$, aimed to acquire the skills required for a target job position $G$. We focus on three key output items: identified skills, learning path and learning content, measured by different metrics that impact on learning experience.



\begin{table}[t]
\centering
\caption{Evaluation results on goal-to-skill mapping.}
\vspace{-10pt}
\setlength{\tabcolsep}{2.3pt}
\begin{tabular}{cl|ccc}
\toprule
 &  & \textbf{Recall} & \textbf{Precision} & \textbf{Goal Alignment} \\
\midrule

\multirow{4}{*}{GPT4o} & DirPrompt & 0.42 & 0.31 & 3.45 \\ 
                       & CoTPrompt & 0.48 & 0.39 & 3.51 \\
\cline{2-5}
                       & \textbf{GenMentor} & \textbf{0.67} & 0.63 & \textbf{4.28} \\
                       & \textit{ w/o Tracks} & 0.63 & \textbf{0.67} & 4.05 \\
 
\midrule

\multirow{4}{*}{Llama} & DirPrompt & 0.37 & 0.35 & 3.18 \\  
                       & CoTPrompt & 0.45 & 0.38 & 3.24 \\
\cline{2-5}
                       & \textbf{GenMentor} & \textbf{0.63} & \textbf{0.61} & \textbf{4.14}  \\
                       & \textit{ w/o Tracks} & 0.61 & 0.58 & 4.01 \\
\bottomrule
\end{tabular}%
\label{tab:results-goal-to-skill-mapping}
\vspace{-12pt}
\end{table}

\subsubsection{Evaluating Goal-to-skill Mapping} 
We compare GenMentor with three approaches: (1) Direct Prompt (DirPrompt) and (2) CoTPrompt, where the CoT reasoning is integrated into the prompt no or yes.
Lastly, we include a variant of GenMentor, (3) w/o Tracks, which removes track-based guidance to assess its impact. All methods use the same learner information and learning goal as input.
To evaluate the quality of goal-to-skill mapping, we regard the skill requirements extracted from job postings as ground truth and use LLM to measure three metrics: Recall, the proportion of identified skills out of ground truth, assessing the comprehensiveness; Precision, the necessity by calculating the proportion of correct skills among the generated outputs; and Goal Alignment, indicating how well the identified skills align with the given learning goal.

As shown in Table ~\ref{tab:results-goal-to-skill-mapping}, GenMentor with GPT-4o outperforms all baselines across all metrics, achieving a Recall of 0.67, Precision of 0.63, and Goal Alignment of 4.28, demonstrating its ability to generate comprehensive, accurate, and goal-aligned skill mappings. The track-based guidance in GenMentor is useful in most metrics, as removing it reduces Goal Alignment to 4.05, highlighting the importance of structured guidance. While CoTPrompt, with its reasoning steps, performs better than DirPrompt, both baselines fall short of GenMentor, particularly in precision and alignment. These results underscore the effectiveness of GenMentor's CoT-enabled fine-tuning method for optimal goal-to-skill mapping.

\subsubsection{Evaluating Learning Path}
To assess the quality of the learning path scheduled by GenMentor, we evaluate its effectiveness in two dimensions: Progression, which measures the logical flow and scalability of difficulty, and Engagement, which assesses the degree to which the path keeps learners motivated and interested. Higher scores on these metrics indicate better performance. We compare GenMentor with DirPrompt and CoTPrompt, introduced above. They are given the same learner profile $U_0$ and skill gap $\Delta S_0$ initialized by GenMentor to schedule the learning path statically.


As shown in Table~\ref{tab:results-learning-path}, GenMentor achieves the highest scores across both metrics, outperforming the baselines in both Progression (4.56) and Engagement (4.71) with GPT-4o, and 4.09 and 4.32 with Llama, respectively. CoTPrompt performs moderately well but remains below GenMentor, while DirPrompt shows significantly lower performance.
It demonstrates that the use of CoT prompts notable benefits for this planning task requiring effective reasoning.
Comparing CoTPrompt without mimicked feedback reveals,  we observe that the \textit{learner simulator} plays a crucial role in further enhancing the quality of the scheduled learning path. These results underscore the importance of incorporating CoT reasoning and mimicked feedback to optimize learning path scheduling.

\begin{table}[t]
\centering
\caption{Evaluation results on Learning Path.}
\vspace{-10pt}
\begin{tabular}{cl|ccc}
\toprule
 &  & \textbf{Progression} & \textbf{Engagement} & \\
\midrule

\multirow{4}{*}{GPT4o} & DirPrompt & 3.95 & 3.80 \\ 
                       & CoTPrompt & 4.38 & 4.63\\
\cline{2-5}
                       & \textbf{GenMentor} & \textbf{4.56} &  \textbf{4.71} \\
\midrule

\multirow{4}{*}{Llama} & DirPrompt & 3.94  & 3.71 \\  
                       & CoTPrompt & 4.07 & 4.17 \\
\cline{2-5}
                       & \textbf{GenMentor} & \textbf{4.09} & \textbf{4.32}  \\

\bottomrule
\end{tabular}%
\label{tab:results-learning-path}
\vspace{-12pt}
\end{table}

\subsubsection{Evaluating Learning Content} 
We compare GenMentor with three baselines: (1) DirGen, which generates content directly without structured guidance; (2) RAG, which incorporates RAG via the search tool; and (3) OutlineRAG, which first prepares a document outline and then drafts each section for integration. Additionally, we consider a variation of GenMentor for comparison: (4) w/o Refinement that skips the refinement step, using initial section drafts as the final document.
To assess the quality of learning content generated by GenMentor, we evaluate its performance across four key dimensions: Goal Relevance, Content Quality, Engagement, and Personalization.
Given the same initialized learner profile $U_0$, skill gap $\Delta S_0$ and learning path $L_0$ derived by GenMentor in the previous stage as input data, these methods produce a set of learning documents of each learning session.

\begin{figure}
    \centering
    \begin{subfigure}[b]{0.43\textwidth}
        \centering
        \includegraphics[width=\textwidth]{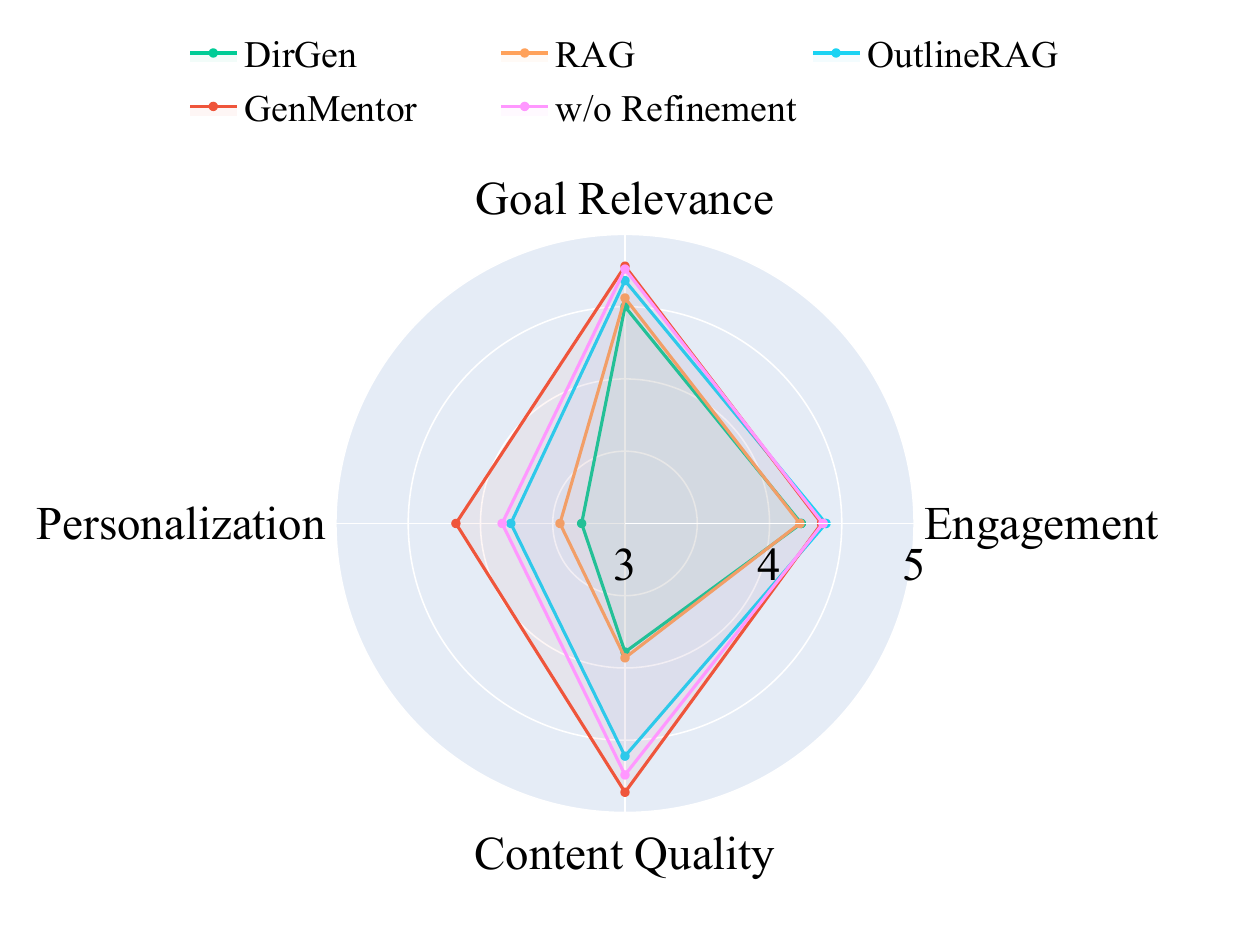}
        \vspace{-10pt}
        \caption{GPT4o}
        \label{fig:first-subtitle}
    \end{subfigure}
    \hfill
    \begin{subfigure}[b]{0.43\textwidth}
        \centering
        \includegraphics[width=\textwidth]{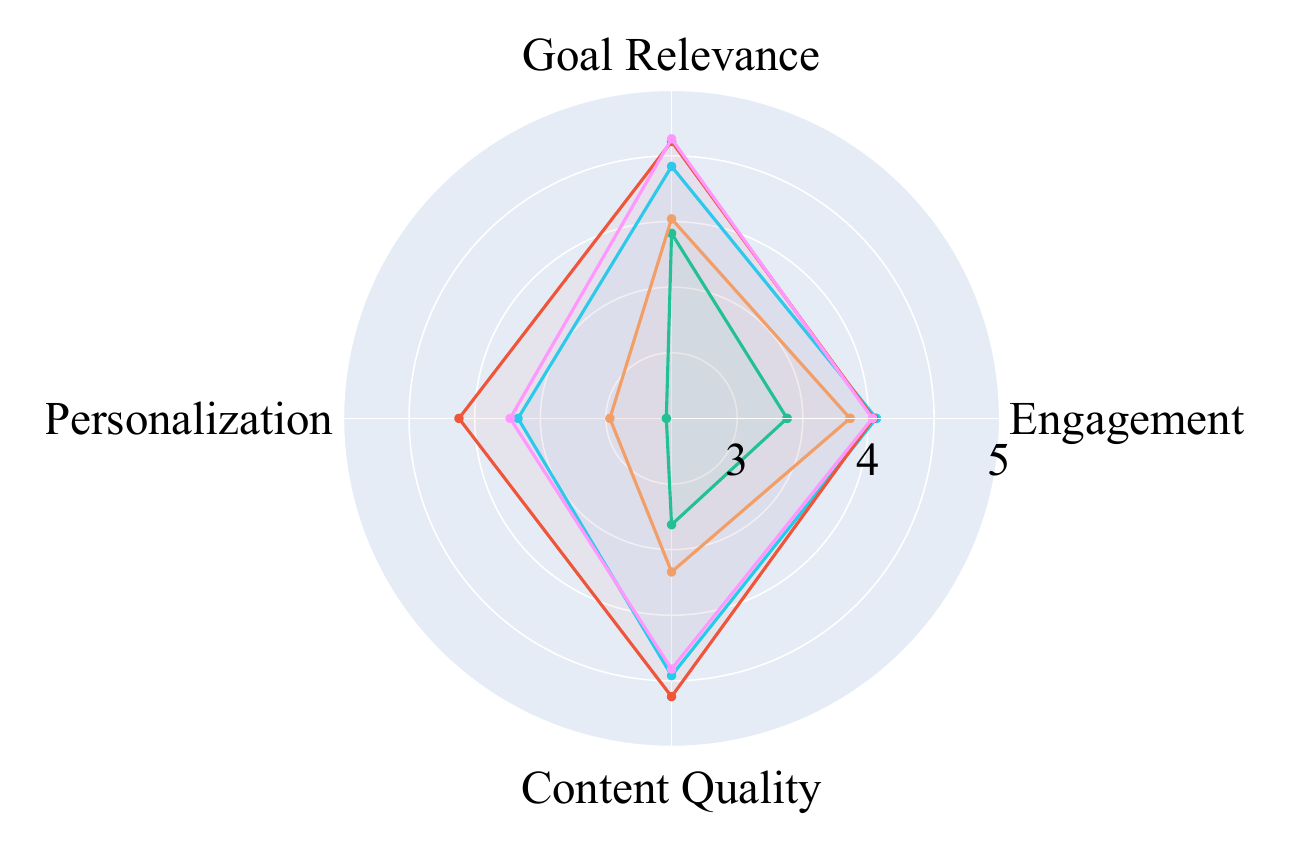}
        \vspace{-10pt}
        \caption{Llama}
        \label{fig:second-subtitle}
    \end{subfigure}
    \vspace{-10pt}
    \caption{Evaluation results on created learning content.}
    \label{fig:content-evaluation}
\vspace{-10pt}
\end{figure}

As shown in Figure~\ref{fig:content-evaluation}, GenMentor consistently outperforms the baselines across most metrics for both GPT-4o and LLaMA-based models, particularly excelling in Personalization. With GPT-4o, GenMentor achieves notable scores in Personalization (4.17) and Content Quality (4.86), showcasing its ability to deliver high-quality content tailored to learners' preferences.
While OutlineRAG and the w/o Refinement perform closely, they fall short in Personalization (3.79 and 3.85), indicating that the refinement step using mimicked learner feedback enhances customization. 
DirGen and RAG exhibit lower performance in multiple dimensions, which highlights the importance of a structured approach in delivering goal-relevant and high-quality content
With Llama-based models, GenMentor maintains its superior performance in Personalization (4.12) and Content Quality (4.62). OutlineRAG continues to deliver moderate results but is significantly outpaced by GenMentor in personalization (3.67). These findings demonstrate GenMentor's ability to align learning content with learner goals and profiles, offering a more personalized, learning experience.

\subsubsection{Human Validation on Automated Evaluation.} To assess the quality of these results in automated evaluation, we compare automated scores with human grading on sampled results. The results show 5 out of 7 metrics exhibit a statistically significant positive correlation. Please see Appendix~\ref{appendix:human-validation} for details.

\subsection{Human Preference Evaluation}
\begin{figure}
    \centering
    \includegraphics[width=0.36\textwidth]{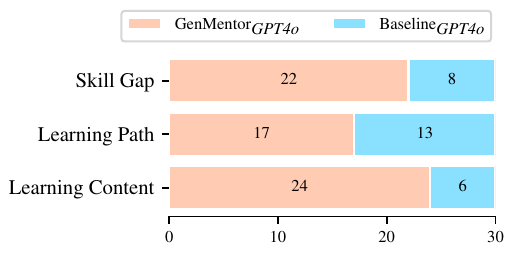}
    \vspace{-12pt}
    \caption{Human comparative preference.}
    \vspace{-15pt}
    \label{fig:human-preference}
\end{figure}

To further evaluate the effectiveness of GenMentor, we conduct a pairwise human evaluation comparing it to strong baselines. For skill gap and learning path, the baseline method is CoTPrompt, while for learning content, the baseline used is OutlineRAG. The results are shown Table~\ref{fig:human-preference}. We observe that GenMentor
was more favored, showcasing its ability to produce high-quality outputs. See Appendix~\ref{appendix:human-comparative-evaluation} for detailed experimental setup and result analysis.

\section{End-to-end Human Study}

We have deployed GenMentor in practice and conducted the human study with professional learners for further evaluation.

\begin{figure*}[t]
    \centering
    \includegraphics[width=0.96\textwidth]{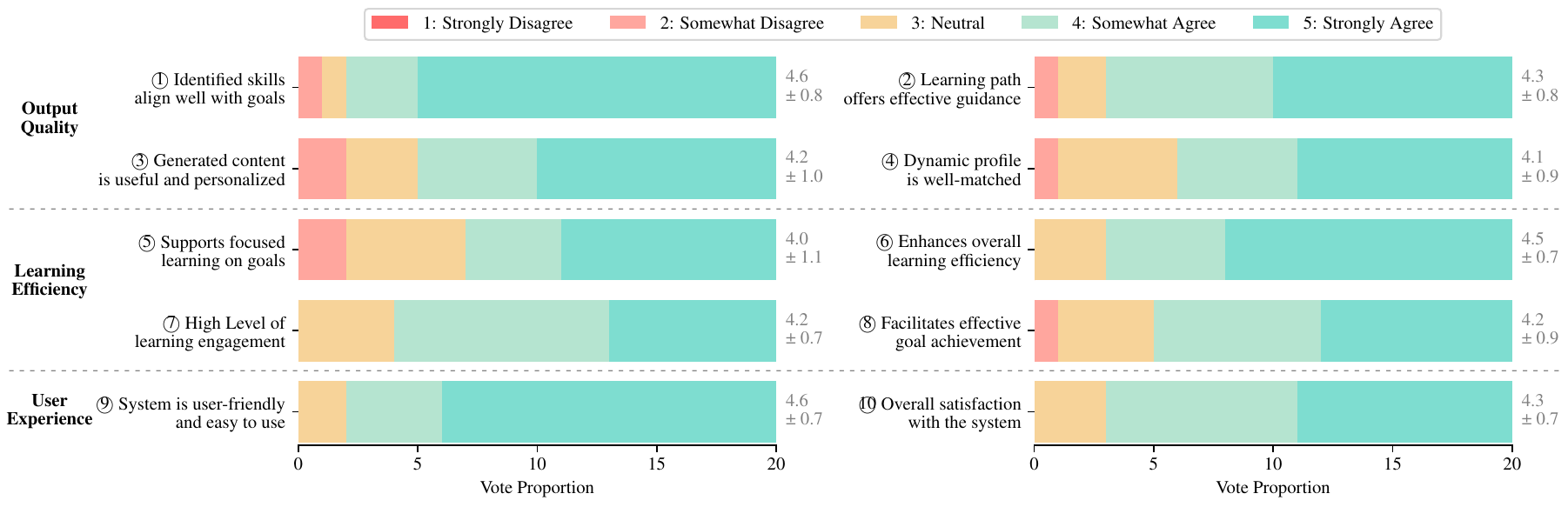}
    \vspace{-8pt}
    \caption{Questionnaire results from 20 participants (questions shortened for clarity). Gray texts are means and std. deviations.}
    \vspace{-10pt}
    \label{fig:survey-results}
\end{figure*}

\subsection{Practical Deployment}
We have deployed GenMentor within our private product, the AIEP platform, launched in October 2024. AIEP is designed to empower employees with AI-driven tools to enhance productivity and streamline skill development. In AIEP, GenMentor supports critical functionalities, including learning path scheduling, resource generation, and learner modeling management, all integrated through API-based interactions. Additionally, we have implemented GenMentor as an independent web-based application tailored for goal-oriented learning and personalized tutoring. This application enables learners to engage in various educational activities for goal achievement, such as skill gap identification, scheduling learning paths, and personalized content delivery. The user-friendly interface facilitates a highly interactive learning experience, allowing learners to achieve their goals efficiently and effectively. This application also has been adopted by employees at Microsoft and partner vendor companies, demonstrating its practical value in enhancing professional development. Please see Appendix~\ref{appendix:web-application} for details on the application.

\subsection{Human Study Procedure}
To study GenMentor's practical effectiveness, we engaged 20 employees from diverse professional backgrounds, including 10 tech-related professionals (e.g., engineers, researchers) and 10 non-tech professionals (e.g., product managers, human resource specialists).
Each participant had prior experience with our GenMentor application as well as traditional MOOC platforms and LLM-based chatbots (e.g., Microsoft Copilot, ChatGPT).
The study included a questionnaire assessing output quality, learning efficiency, and user experience, followed by interviews to discuss GenMentor's strengths and limitations compared to existing tutoring systems. 

\subsection{Questionnaire Findings}

Results are shown in the table~\ref{fig:survey-results} and our findings are as follows.

\textit{Clear learning guidance for goal achievement} (\textcircled{1}\textcircled{2}). 18 participants agreed that the system effectively identifies skills aligned with their goals, resulting in a high rating of 4.6 ± 0.8. The learning path offered by GenMentor was rated as 4.3 ± 0.8, with participants noting that it breaks down learning objectives into manageable steps. One participant highlighted, “The clear guidance made it easier to navigate addressing complex tasks and achieve goals.”

\textit{Personalized and contextually relevant content} (\textcircled{3}\textcircled{4}).
GenMentor's personalized content was well-received, with participants rating the generated content as useful and personalized, scoring 4.2 ± 1.0. However, the dynamic profile matching received a slightly lower rating of 4.1 ± 0.9, suggesting that there is room for improvement in refining the learning modeling approach. Participants commented on the generated learning materials, with one stating, “The content is well-structured and targeted, helping me stay focused. I find my name in the personalized content surprisingly.”

\textit{Improved goal-oriented learning experience} (\textcircled{5}-\textcircled{8}). Participants reported significant improvements in learning efficiency, with more than 80\% noting enhanced efficiency and high-level engagement in learning. Additionally, more than 14 participants agreed that GenMentor made them focused and facilitated goal achievement. One participant stated, "The system kept me engaged and motivated by adapting to my needs and providing timely feedback."

\textit{Intuitive and User-friendly System Design} (\textcircled{9}-\textcircled{10}).
The user-friendly design of GenMentor is a supportive point in utility, with participants rating the system's ease of use at 4.6 ± 0.7. The overall satisfaction with the system also received a commendable score of 4.3 ± 0.7, with participants appreciating the seamless interaction and clear layout. A participant noted, “The interface is intuitive and easy to navigate, making the learning experience enjoyable.”

\subsection{Interview and Discussion}
To gain deeper insights into GenMentor's potential, we conduct interviews comparing it with existing learning platforms and tutors. 

\subsubsection{Comparison with Traditional MOOCs.}
Compared to MOOC platforms, participants valued GenMentor for its ability to customize learning paths, narrow the learning focus, and enhance content personalization. 15 participants specifically highlighted its automated skill gap identification feature, which helps clarify learning needs and streamline content delivery. Additionally, the platform's dynamic content adjustment and progress tracking capabilities were praised for keeping learners engaged and facilitating efficient goal achievement (highlighted by 8 participants).
Despite these strengths, participants suggested areas for improvement, such as incorporating more diverse and interactive content formats (noted by 13 participants) and enhancing the depth of subject matter (noted by 12 participants). Furthermore, Participants identified scenarios where GenMentor excels, such as providing personalized guidance for specific skill development (highlighted by 15 participants) (noted by 13 participants). 
While MOOCs offer broad-topic coverage through systematic general curricula, GenMentor's tailored and adaptive approach clearly outperforms MOOCs in scenarios requiring focused, goal-oriented learning.





\subsubsection{Comparison with Search-enhanced Chatbots}
Participants identified GenMentor's key advantages over search-enhanced chatbots in providing personalized learning paths and interactive guidance for task-specific learning (highlighted by 17 participants). 13 participants praised its clear and structured learning progression, which uniquely aligns content with learners' goals. Unlike the generic and reactive responses of chatbots, GenMentor reduces proactively asking of learners and fostering focused learning, while dynamically adjusting its content based on real-time learner progress (noted by 9 participants).
While GenMentor outperformed chatbots in goal-oriented learning experience, participants suggested improvements, such as adding more diverse multimedia resources (noted by 8 participants) and optimizing system responsiveness to enhance real-time interactions (noted by 7 participants). Participants found GenMentor particularly advantageous for scenarios like breaking down complex topics into manageable learning paths (noted by 16 participants). Although chatbots may excel in quick problem-solving, GenMentor's structured, goal-driven approach delivers a deeper, more interactive learning experience.

%% file: sections/conclusion.tex
\section{Conclusion}
In this paper, we presented GenMentor, an LLM-powered multi-agent framework for goal-oriented learning in ITS, designed to deliver highly personalized, goal-oriented learning experiences. GenMentor advances beyond existing ITSs by proactively guiding learners efficiently toward their goals through accurate skill gap identification, adaptive learner profiling, and personalized resource delivery.
Through extensive evaluations, including automated and human evaluation, GenMentor demonstrated superior performance in key outputs, including identifying skill gaps, aligning learning paths, and delivering tailored content. Real-world deployment and user studies further validated its effectiveness in fostering efficient and personalized learning in professional contexts. 
Overall, with its adaptive design and learner-centric approach,  GenMentor showcases the transformative potential of LLMs in advancing personalized and goal-oriented education.

%% file: sections/appendix.tex
\appendix
\section{Appendix}
Due to page limitation, we provide supplementary resources at \href{https://github.com/GeminiLight/gen-mentor}{https://github.com/GeminiLight/gen-mentor} in addition to the Appendix, including demo, prompts, data and more insights.

\begin{figure*}
    \centering
    \includegraphics[width=0.88\textwidth]{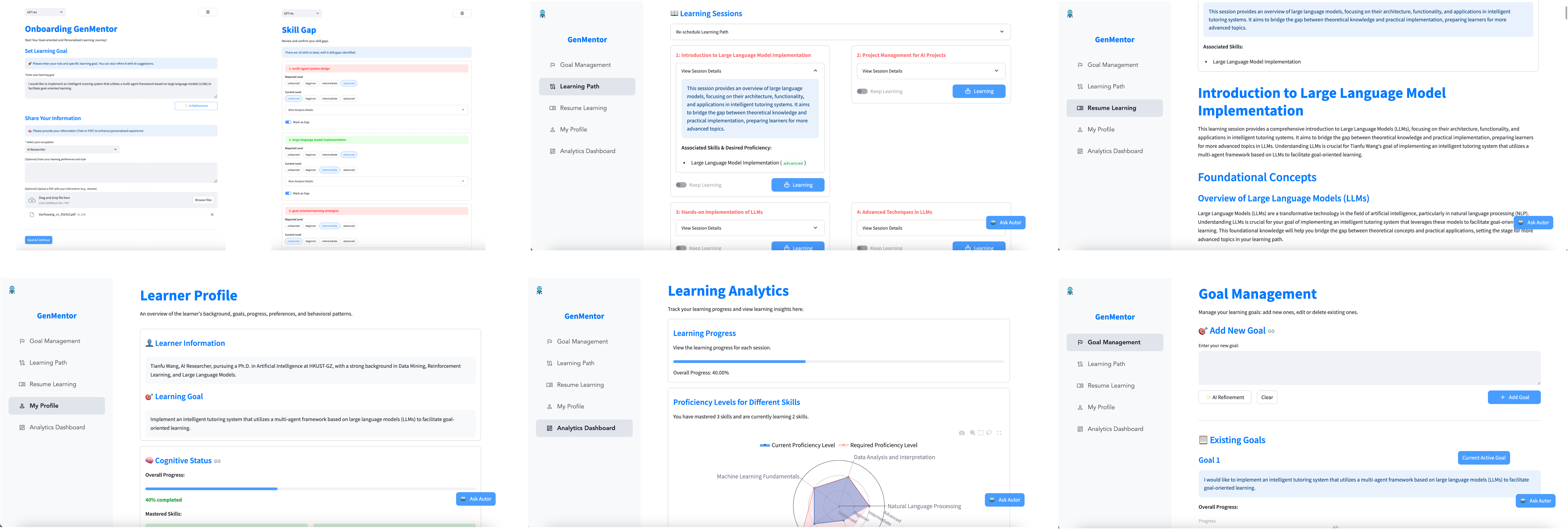}
\vspace{-10pt}
    \caption{The core interfaces of GenMentor's web application. Please see our supplementary resources for more details.}
    \label{fig:web-demo}
\vspace{-10pt}
\end{figure*}

\subsection{Details on Goal-to-skill Dataset}
\label{dataset-details}

To construct the goal-to-skill dataset, we use the LinkedIn job posting dataset from Kaggle\footnote{https://www.kaggle.com/datasets/arshkon/linkedin-job-postings}, which contains more than 0.12 million job postings across various positions. We filter the dataset based on the word count, retaining only postings with at least 500 words.
This process results in a refined subset of 58,064 postings. From this subset, we randomly sample 10,000 postings to create the training dataset and 200 postings as the validation dataset, ensuring balanced representation across position types.
For each sample, we employ GPT4o to extract the job summaries paired with their corresponding skill lists. Regarding the extracted skill lists as output, we use the CoT-enabled completion method to incorporate the immediate reasoning steps. This process involves breaking down the goals into key duties, identifying the required skills for each duty, and determining the proficiency levels needed, resulting in samples of <job summary, reasoning tracks, required skills>. 
This curated dataset provides clear mappings from goals (job objectives) to skills (necessary competencies), used for fine-tuning \textit{gap identifier} and evaluating outputs of goal-to-skill mapping. 
Examples of these samples are available in the supplementary resources.

\subsection{Human Validation on LLM Scoring}
\label{appendix:human-validation}
To validate the quality of the LLM-based automated evaluation results, we conduct an experiment comparing automated scores with human grading for GenMentor's output. Concretely, we invite two experts with extensive experience in Python development to participate in this validation. 
For each output item (i.e., goal-to-skill mapping, learning path, and learning content), 20 data points, all related to the occupation type of Python developer, were randomly sampled from the automated evaluations. Each data point represented the automated evaluation score of one method's 
Two independent evaluators were tasked with providing human grading for the same data points, and their average scores were used as the human judgment benchmark. We calculate the Pearson correlation between the automated evaluation scores and the human grading scores to assess the alignment between the two scores. 
As depicted in Table~\ref{tab:human-validation}, the results show that 5 out of 7 evaluation metrics exhibit a statistically significant positive correlation between automated scores and human grading. This indicates the relative consistency of the LLM-based automated evaluation method and human sense. 

\begin{table}[t]
\centering
\setlength{\tabcolsep}{2pt}
\caption{Pearson correlation between two types of scores.}
\vspace{-10pt}
\renewcommand{\arraystretch}{1.1}
\begin{tabular}{cc|cc}
\toprule
\textbf{Category} & \textbf{Metric} & \textbf{Correlation} & \textbf{p-value} \\ 
\midrule
\multirow{1}{*}{\centering Goal2Skill Mapping} 
    & Goal Alignment    & 0.51 & < $4^{-2}$ \\ 
\hline
\multirow{2}{*}{\centering Learning Path} 
    & Progression        & 0.47 & < $2^{-2}$ \\ 
    & Engagement   & 0.39 & \underline{< $3^{-1}$}   \\ 
\hline
\multirow{4}{*}{\centering Learning Content} 
    & Content Quality   & 0.52 & < $1^{-2}$ \\ 
    & Goal Relevance    & 0.46 & < $1^{-2}$ \\ 
    & Engagement        & 0.38 & < $4^{-2}$ \\ 
    & Personalization   & 0.42 & \underline{< $8^{-2}$} \\ 
\bottomrule
\end{tabular}
\label{tab:human-validation}
\vspace{-14pt}
\end{table}

\subsection{Human Preference Evaluation Details}
\label{appendix:human-comparative-evaluation}
\subsubsection{Experiment Setup}
We invited 5 participants (including 3 software engineers and 2 product managers) to set learning goals related to occupations. Participants provided their resumes as learner information, which, along with each learning goal, served as input for both GenMentor and the baseline method. Each participant repeated this process 6 times for the distinct learning goals. The evaluation followed a step-by-step process. Both systems simultaneously generated skill gaps based on the provided information and learning goal. Using the preferred skill gap as additional input, two methods were used to generate corresponding learning paths. Finally, learning content was produced by both methods based on the same learning path. At each stage, participants reviewed the outputs side-by-side and selected their preferred option.

\subsubsection{Results and Analysis}
As illustrated in Figure~\ref{fig:human-preference}, results reveal that GenMentor outperformed the baseline systems across all evaluation items, further showing the effectiveness of the automated scores. For skill gap identification, GenMentor was preferred in 22 out of 30 cases, demonstrating its superior ability to align skills with learning goals compared to CoTPrompt. For learning path scheduling, GenMentor was chosen in 17 cases, reflecting its effectiveness in generating goal-oriented paths. For learning content generation, GenMentor was favored in 80\% of cases, showcasing its ability to produce high-quality and personalized materials.

\subsection{Web-based Application of GenMentor}
\label{appendix:web-application}
To illustrate the practical utility of GenMentor, we provide key interfaces of its web application in Figure~\ref{fig:web-demo}. Learners begin on the onboarding page by setting their learning goals and providing their background information. GenMentor then identifies skill gaps, which learners can review and confirm. Based on the confirmed gaps, a personalized learning path is scheduled, with options for learners to manually adjust or allow automated refinements based on their updated profiles.
During each learning session, tailored content is delivered to learners, ensuring alignment with their goals. Learners can also review and update their profiles on the learner profile page, maintaining control over their learning journey. This streamlined design underscores GenMentor's ability to deliver adaptive and goal-oriented learning experiences.